\documentclass[10pt,twocolumn,letterpaper]{article}

\usepackage{cvpr}
\usepackage{times}
\usepackage{epsfig}
\usepackage{graphicx}
\usepackage{amsmath}
\usepackage{amssymb}
\usepackage{epstopdf}
\usepackage{xcolor}
\usepackage{subcaption}
\usepackage[ruled,vlined]{algorithm2e}

\usepackage{colortbl}
\definecolor{mygray}{gray}{.9}
\definecolor{mypink}{rgb}{.99,.91,.95}
\usepackage[pagebackref=true,breaklinks=true,letterpaper=true,colorlinks,bookmarks=false]{hyperref}

\usepackage[breaklinks=true,bookmarks=false]{hyperref}

\cvprfinalcopy 

\def\cvprPaperID{5712} 

\begin{document}

\title{Domain-Aware Dynamic Networks}

\author{Tianyuan Zhang\\
Peking University\\
{\tt\small tianyuanzhang@pku.edu.cn}
\and
Bichen Wu\\
University of California, Berkeley\\
{\tt\small bichen@berkeley.edu}
\and
Xin Wang\\
University of California, Berkeley\\
{\tt\small xinw@berkeley.edu}
\and
Joseph Gonzalez\\
University of California, Berkeley\\
{\tt\small jegonzal@berkeley.edu}
\and
Kurt Keutzer\\
University of California, Berkeley\\
{\tt\small keutzer@berkeley.edu}
}

\maketitle

\begin{abstract}
    Deep neural networks with more parameters and FLOPs have higher capacity and generalize better to diverse domains. But to be deployed on edge devices, the model's complexity has to be constrained due to limited compute resource. In this work, we propose a method to improve the model capacity without increasing inference-time complexity. Our method is based on an assumption of data locality: for an edge device, within a short period of time, the input data to the device are sampled from a single domain with relatively low diversity. Therefore, it is possible to utilize a specialized, low-complexity model to achieve good performance in that input domain. To leverage this, we propose Domain-aware Dynamic Network (DDN), which is a high-capacity dynamic network in which each layer contains multiple weights. During inference, based on the input domain, DDN dynamically combines those weights into one single weight that specializes in the given domain. This way, DDN can keep the inference-time complexity low but still maintain a high capacity. Experiments show that without increasing the parameters, FLOPs, and actual latency, DDN achieves up to 2.6\% higher AP50 than a static network on the BDD100K object-detection benchmark.
\end{abstract}

\section{Introduction}







\begin{figure}
    \centering
    \includegraphics[width=\linewidth]{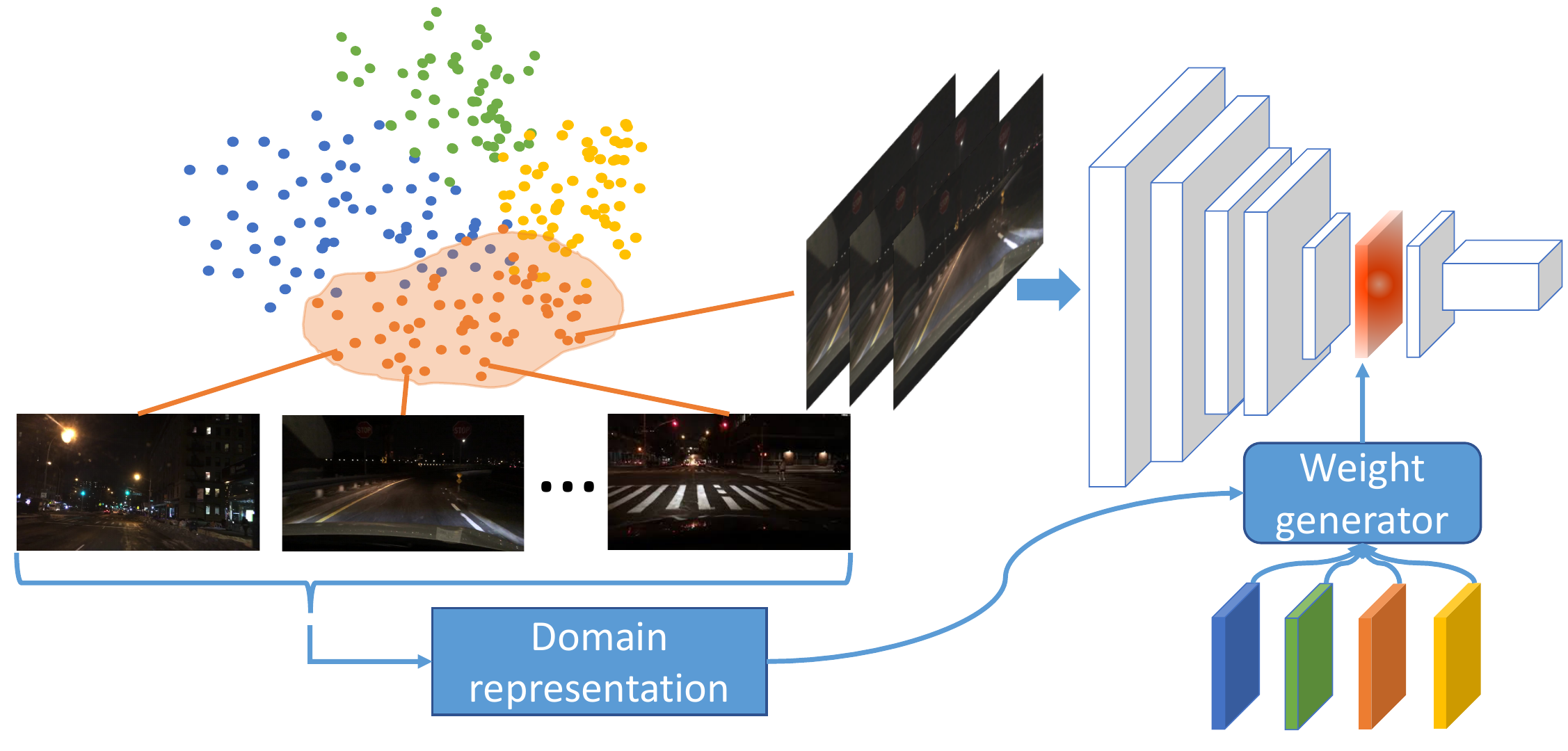}
    \caption{Overview of domain-aware dynamic network. The colored dots represent data drawn from different domains. During the inference, we collect input images to compute the domain representation, which are then fed into a controller to determine how to combine weights of experts into 
    a static small network specialized for the input domain. Under the data locality assumption, as long as the input domain remains unchanged, the weights of the dynamic network also stay the same, so we can load the network parameters once and re-use them for all the input images. This allows us to improve model capacity without increasing the inference-time complexity. 
    }
    \label{fig:dadn}
\end{figure}

When deploying deep neural networks on the edge, we often face a dilemma between the model capacity and its computational complexity~\cite{wu2018shift,wang2018skipnet,shazeer2017outrageously}. 

\begin{figure}
    \centering
    \includegraphics[width=.8\linewidth]{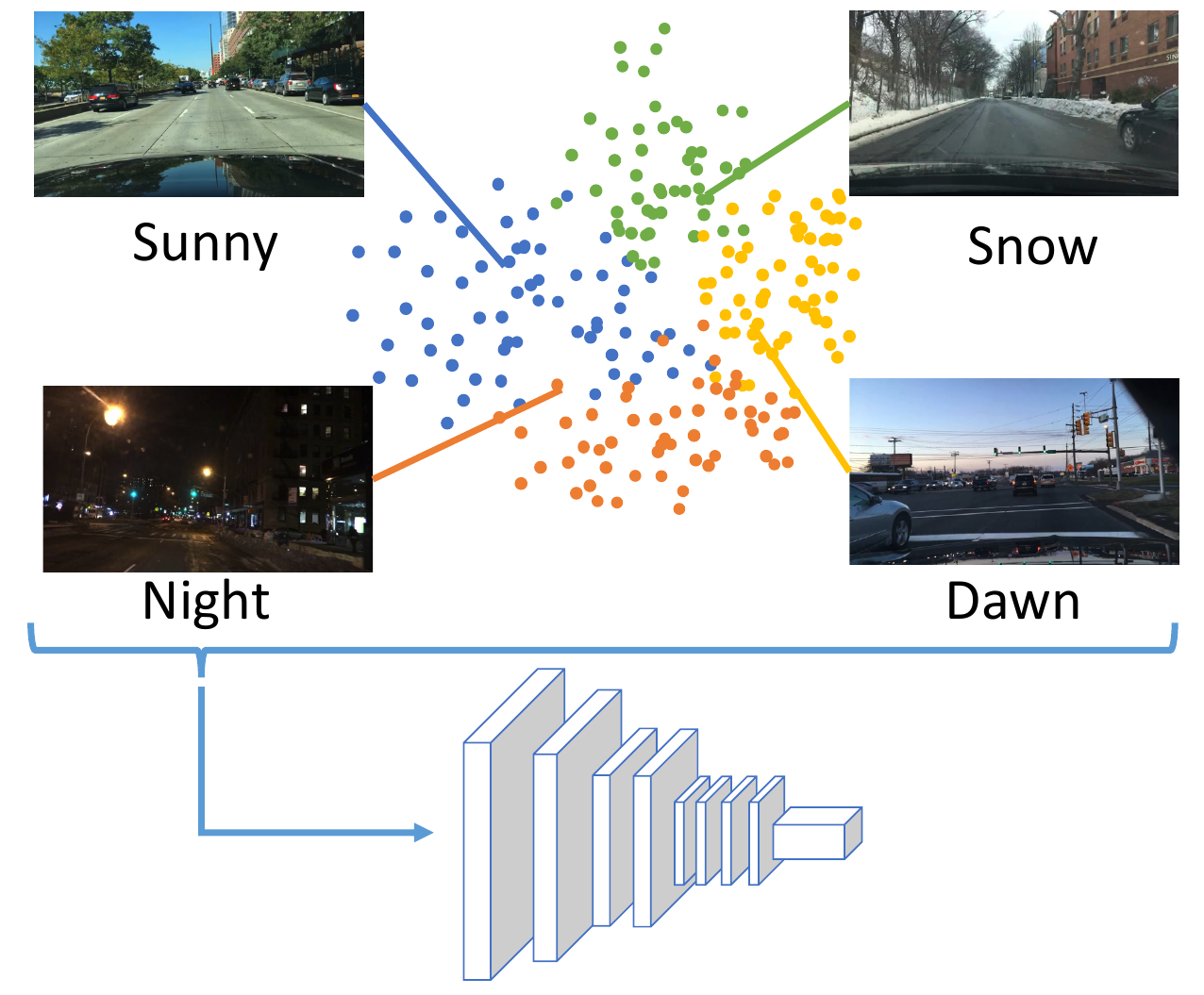}
    \caption{A large and diverse dataset can contain images from a wide variety of domains. In order to improve the model's capacity and its generalizability across domains, the conventional approach is to use the dataset to train a large neural network with more parameters and FLOPs. 
    }
    \label{fig:single_model}
\end{figure}

On the one hand, we hope a neural network can achieve high accuracy in a wide variety of scenarios. For example, for autonomous driving, a neural network-based object detection model should accurately detect target objects regardless of weather conditions, time of the day, and so on. In the computer vision and machine learning community, a scenario under the same weather, time of the day, and other conditions can be regarded as a domain. Images sampled from the same domain share many similarities, while images across different domains are more diverse. The conventional way to achieve high accuracy across many domains is to collect a large-scale dataset that contains diverse domains\cite{yu2018bdd100k, shao2019objects365} and use that dataset to train a single gigantic neural network (Figure~\ref{fig:single_model}). Previous work~\cite{mahajan2018exploring, recht2019imagenet} has shown that the larger the model is, in terms of parameters and FLOPs, the better it can generalize across domains.

On the other hand, deploying the model on edge processors requires us to reduce the model complexity. Edge processors have limited computing resources, and many real-world applications, such as autonomous driving, impose strict requirements on the inference speed \cite{wu2017squeezedet}. These two constraints together force us to decrease the model complexity by reducing parameters and FLOPs.  


Thus, we attempt to address the question of whether it is possible to increase a neural network's capacity without increasing its computational complexity? To be more precise, we measure the capacity of a neural network by its accuracy on a large and diverse dataset, and we measure a model's complexity by parameter size and FLOPs. Most of the previous works address this problem by optimizing neural network models \cite{Wu:EECS-2019-120}. This includes efficient network architecture design \cite{ma2018shufflenet, wu2017squeezedet, wu2018squeezeseg, gholami2018squeezenext, howard2017mobilenets, sandler2018mobilenetv2, zhang2018shufflenet, wu2018shift, howard2019searching, yang2019synetgy}, pruning \cite{han2015deep, he2017channel, liu2019metapruning, ding2019approximated}, quantization \cite{dong2019hawq, wang2019haq, wu2018mixed}, and automated neural architecture search \cite{zoph2016neural, tan2019mnasnet, wu2019fbnet, guo2019single, dai2019chamnet}. 

In this work, we approach this problem from a new perspective by exploiting the input data's \textbf{locality}: although a neural network on an edge device should generalize to diverse domains, within a short period of time, the environment around the device stays relatively stable. For example, when an autonomous vehicle drives at night, within the range of minutes to hours, it does not need to deal with day-time images. When this property holds, input images to a neural network are likely to come from a single domain with relatively small diversity. Thus, it is possible to use a low-complexity model specialized to this domain to achieve good performance.

\begin{figure}
    \centering
    \includegraphics[width=.7\linewidth]{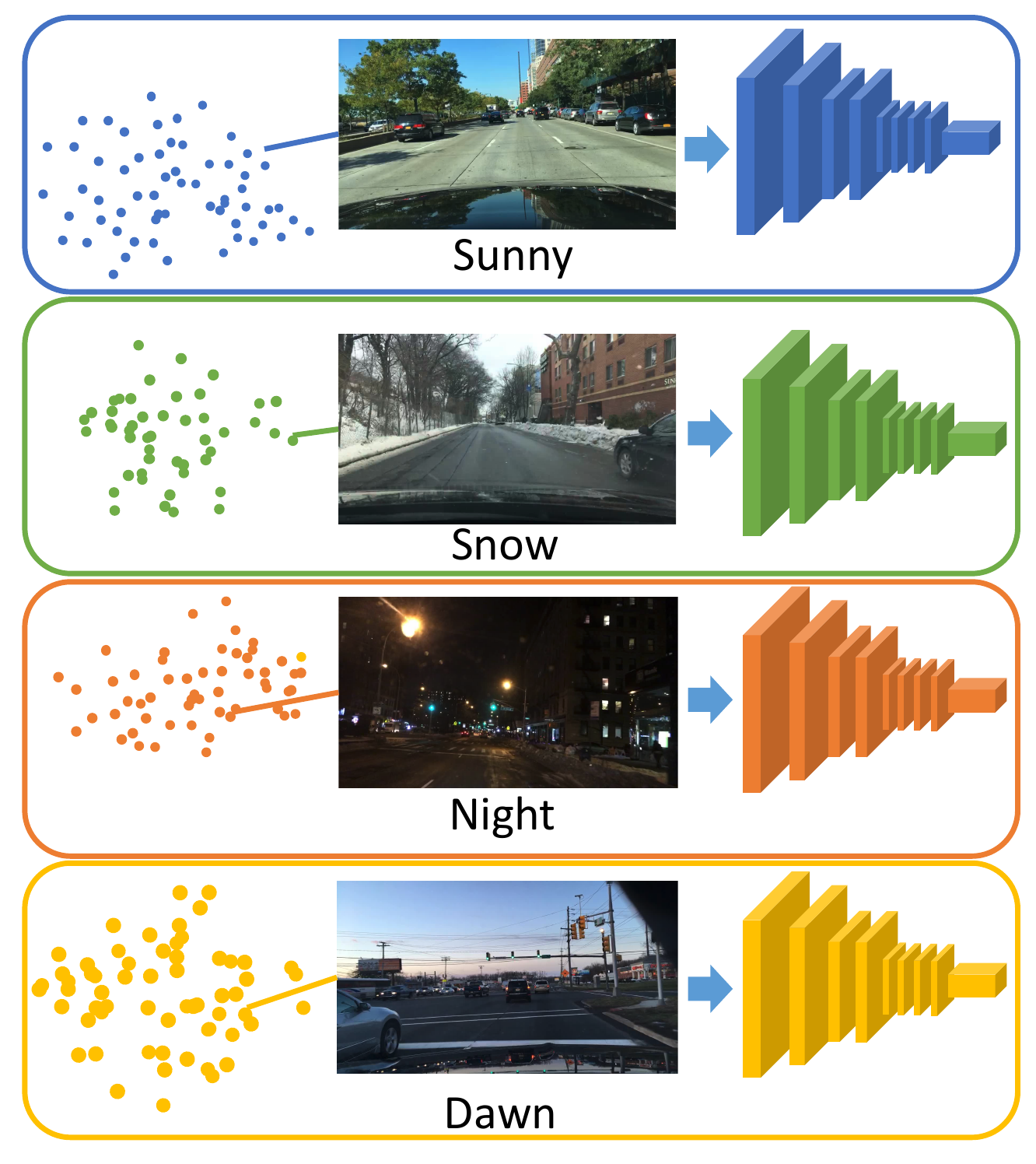}
    \caption{A straightforward idea to utilize the data locality. We divide the entire dataset into domains, and for each domain, we train a small and specialized network. During inference, we only load the specialized small model corresponding to the input domain. 
    }
    \label{fig:model_pool}
\end{figure}

To leverage the locality of input data, in this paper, we propose Domain-aware Dynamic Networks (DDN), as illustrated in Figure \ref{fig:dadn}. 
DDN is a dynamic neural network whose weights are conditioned on the input domain. In a layer of DDN, the actual weight of the convolution is a linear combination of several weights. The factor for each weight is dynamic and conditioned on the input domain. During the inference, we extract a domain representation from input data and use it to compute the weight factors to combine them together. As long as the input domain stays the same, we can reuse the same network for all the input images. As a result, the inference stage complexity of DDN stays the same as as a regular neural network, but DDN can still leverage the extra model parameters to generalize to more than just the current domain. 

Our experiments show that without increasing the inference-time model complexity, DDN achieves up to 2.6\% higher AP50 than single networks and other baselines on the BDD100K dataset.

\section{Related work}

\paragraph{Efficient neural networks} Most prior work on model acceleration focus on either designing a compact light-weight model \cite{ma2018shufflenet, wu2017squeezedet, howard2017mobilenets, sandler2018mobilenetv2, zhang2018shufflenet, wu2018shift, howard2019searching}, or pruning and quantization \cite{han2015deep, he2017channel, dong2019hawq, liu2019metapruning, wang2019haq}. Recently, people also use Neural Architecture Search (NAS) to automatically optimize neural networks \cite{zoph2016neural, tan2019mnasnet, wu2019fbnet, guo2019single, dai2019chamnet, cai2019once}. In this paper, we propose a new dimension to explore the problem by utilizing dynamic networks and the data locality. Our method is orthogonal to the effort of building more efficient neural networks.

\paragraph{Conditional Computation}  
For a dynamic or conditional computation model, the computation graph or the weight of the model are dependent on the input. One application of conditional computation is to reduce the average computational cost by skipping part of the model in an input-dependent fashion \cite{wang2018skipnet,veit2018convolutional, lin2017runtime, bejnordi2019batch, shazeer2017outrageously}.  Another class of  algorithms try to improve a model's in-variance or equivalence to input variations. This include deformable convolution \cite{dai2017deformable, zhu2019deformable} and Spatial Transformer Network \cite{jaderberg2015spatial}. 
Recently Yang el al.\cite{yang2019soft} demonstrated that a simple soft routing algorithm could stabilize the training and improve the performance and inference cost trade-off. Most of the previous work on conditional computation are sample-dependent such that the computational graph or weights changes per input sample. This is not hardware friendly as it makes the scheduling of computation more complicated. In our work, we propose a domain-dependent network, so network remains static as long as the input domain does not change. This is more hardware-friendly than previous work.  TAFE-Net~\cite{wang2019tafe} designed for low-shot recognition generates weights conditioned on the classification tasks while we in this work focus on different sub-domains of the data.

\paragraph{Adaptation} The locality of data considered in this work is related to \textit{continuous adaptation}, which is a variant of \textit{continual learning}. Continuous adaptation attempts to solve a nonstationary task by adapting the learning algorithm during execution gradually.  The difference between our algorithms and popular continual learning \cite{al2017continuous, li2017learning} and adaptation methods \cite{finn2017model, sun2019test} is that our model do not ``learn'' to adapt to new data. Instead, it extracts features from new data and adapt to the new domain in a feed-forward manner. Also, our work is related to domain adaptation \cite{zhao2018unsupervised, wu2019squeezesegv2, zhao2019multi, yue2019domain, hoffman2017cycada}. Most of the work in this area focus on designing training recipes and learning techniques while our work focus on improving the model capacity. 


\section{Methodology}
In this section, we introduce the idea of domain-aware dynamic networks (DDN). We first start from formulating a static neural network as 
\begin{equation}
    \hat{y} = f(\mathcal{W}, x)
\end{equation}
where $x$ is the input to the network, $\hat{y}$ is the output, $\mathcal{W} = [W_1; W_2; \cdots W_n]$ denotes the weights of the network, $W_i$ is the weight of the $i$-th layer. At each layer, the computation can be expressed as
\begin{equation}
    z_i = \sigma(W_i * z_{i-1})
\end{equation}
where $z_i$ is the intermediate output of layer-$i$ and $\sigma(\cdot)$ is a non-linear activation function. $*$ denotes a matrix-vector multiplication if the network is a multi-layer perceptron, or a convolution if it is a convolutio neural network. 
For a given dataset $\mathcal{D} = \{(x, y)\}$, where $x$ represents input data and $y$ represents labels, we train a static neural network to fit the dataset as
\begin{equation}
    \underset{\mathcal{W}}{\text{min}} ~ \sum_{(x, y) \in \mathcal{D} } \mathcal{L} (f(\mathcal{W}, x), y). 
    \label{eqn:single_model}
\end{equation}
$\mathcal{L}(\cdot, \cdot)$ is the loss function between the prediction and the label. The dataset $\mathcal{D}$ may contain samples from many diverse domains, making it difficult to fit using only a small model. To improve the performance, the conventional way is simply to increase the model capacity by adding more parameters and FLOPs. However, due to the limited compute resource available on an edge device, using more complex models is not always feasible. 

\subsection{The locality of data}
In many applications such as autonomous driving and security cameras, a neural network usually operates on a stream of continuous input data. Within a short period of time, the environment around an edge device tends to remain relatively stable. For example, an autonomous vehicle that drives at night may not need to deal with day-time images within a range of a minutes to hours. Because of this, the input data to a neural network have relatively lower diversity. We call this property the ``locality of data''. 

More formally, for a large and diverse dataset $\mathcal{D}$, we can divide it into $T$ smaller domains $D_\tau$ such that $\mathcal{D} = \cup_{\tau=1}^T \mathcal{D}_\tau$. The strategy to divide the dataset can be arbitrary, but to leverage the locality of data, we can choose to use environmental attributes that are relatively stable over some period of time. For example, we can partition the dataset by different weather conditions, geo-locations, time-of-the-day, and so on. For each attribute-$i$, we use $\mathcal{A}_i$ to represent a set of possible values of attribute-$i$, so the number of domains is $T = \prod_i \#(\mathcal{A}_i)$, where $\mathcal{A}_i$ where $\#(\mathcal{A}_i)$ is the cardinality of $\mathcal{A}_i$. Within each domain $D_\tau$, input data have significantly lower diversity, as illustrated in Figure~\ref{fig:bdd-images}. So it is possible for us to train a small network to specialize in one domain. During inference, as long as the environment remains stable, we can use the same small network to process all the input. 

\begin{figure}[ht]
\centering
\begin{tabular}{c|c}
\includegraphics[width=.45\linewidth]{Images/bdd-images/day-clear-highway-1.jpg}     &
\includegraphics[width=.45\linewidth]{Images/bdd-images/night-clear-city-1.jpg}
\\
\includegraphics[width=.45\linewidth]{Images/bdd-images/day-clear-highway-2.jpg}

&
\includegraphics[width=.45\linewidth]{Images/bdd-images/night-clear-city-2.jpg}
\\

\end{tabular}

\caption{Images from BDD100K validation set. Images on the left are sampled from a domain with attributes: ``daytime''-''highway''. Images on the right are with attributes: ``night''-``city street''. We can clearly see that the intra-domain variation is much smaller than the cross-domain variations. }
\label{fig:bdd-images}
\end{figure}





\subsection{A straightforward idea: a pool of specialized models}
\label{sec:model_pool}

To leverage the locality of data, a straightforward idea is to divide and conquer. Instead of building one gigantic model, we build a pool of small and specialized models that each one corresponds to a domain $\mathcal{D_\tau}$, as illustrated in Figure \ref{fig:model_pool}. Formally, instead of using one neural network to fit the entire dataset as in equation (\ref{eqn:single_model}), for each domain $\mathcal{D_\tau}$, we train a small neural network to fit the the specific domain.  
\begin{equation}
    \underset{  \mathcal{W_\tau }}{\text{min}} ~ \sum_{(x, y) \in \mathcal{D_\tau} } \mathcal{L} (f_\tau (\mathcal{W_\tau}, x), y). 
\end{equation}
In practice, however, although $\mathcal{D_\tau}$ has lower diversity, it also contains significantly fewer samples which is insufficient to train a model. So we first train the network on the entire dataset $\mathcal{D}$, and use the solution of equation (\ref{eqn:single_model}) to initialize the model and finetune it on $\mathcal{D_\tau}$. Eventually, we obtain a pool of $T$ models $\{f_\tau(\mathcal{W_\tau}, x)\}_{\tau=1}^T$. During inference, for a given input domain $D_\tau$, we just need to load the corresponding model to perform inference at low cost. 

However, this solution has several problems. 
First, the number of domains grows exponentially with the number of attributes used to divide $\mathcal{D}$. When $T$ is large, it becomes infeasible to store all the models on the edge device.  
Second, as $T$ grows, the number of samples available in each domain $\mathcal{D_\tau}$ drops significantly, making it difficult to train a model or even finetune a model that is pre-trained on $\mathcal{D}$. Also, since each model $f_\tau (\mathcal{W_\tau}, x)$ is only trained on its own domain $\mathcal{D}_\tau$, it loses the chance to learn features that are invariant across different domains. Our experiment confirms that a pool of specialized models does not offer accuracy improvement over a single network.

\subsection{Domain-aware dynamic networks}
To overcome the drawbacks of model pools, we propose domain-aware dynamic networks, as illustrated in Figure \ref{fig:dadn}. A domain-aware dynamic network is essentially a ``supernet'' as in \cite{wu2019fbnet, yang2019soft}. At each layer of the network, the weight of a layer is computed as a linear combination of several weights. The factor for each weight is dependent on the input domain $\mathcal{D}_\tau$, but is static with respect to each individual input $x$. Formally, a domain-aware dynamic network is defined as  
\begin{equation}
    \hat{y} = f(\mathcal{W}(\mathcal{D}_\tau), x)
\end{equation}
where $x \in \mathcal{D}_\tau$ is a sample drawn from domain $\mathcal{D}_\tau$. Note that the weight of the neural network is a function of the domain $\mathcal{D}_\tau$, but it is not dependent on each individual input $x$. Following \cite{yang2019soft}, each layer of the network is defined as 
\begin{equation}
    z_i = \sigma(W_{i}(\mathcal{D}_\tau) * z_{i-1}),
    \label{eqn:layer_func}
\end{equation}
where $*$ denotes a matrix-vector multiplication. The weight of each layer $W_{i}(\mathcal{D}_\tau)$ is computed as 
\begin{equation}
    W_{i}(\mathcal{D}_\tau) = \sum_j \alpha_{i,j}(\mathcal{D}_\tau) \times W_{i,j}.
    \label{eqn:dadn_weight}
\end{equation}
$W_{i,j}$ is the $j$-th weight at the $i$-th layer of the network. $\alpha_{i,j}(\cdot)$ is a scalar function that is dependent on the input domain $\mathcal{D}_\tau$ and is defined as
\begin{equation}
    \alpha_{i,j}(\mathcal{D}_\tau) = c_{i,j}(\theta_{i,j}, \mathcal{E}(\mathcal{D}_\tau))
    \label{eqn:alpha}
\end{equation}
$c_{i,j}(\cdot, \cdot)$ represents a controller function that decides the factor for $W_{i,j}$. $\theta_{i,j}$ is its trainable weight. Specifically, the controller function is composed of one \textit{Sigmoid} function after one linear transformation parameterized by $\theta_{i,j}$. 
$\mathcal{E}(\cdot)$ computes a domain representation vector for $\mathcal{D}_\tau$. There can be many ways to extract the domain representation, but in our experiment, we define $\mathcal{E}(\cdot)$ as 
 \begin{equation}
     \mathcal{E}(\mathcal{D}_\tau) = \frac{1}{|\mathcal{D}_\tau|} \sum_{x\in \mathcal{D}_\tau } e(x),
     \label{eqn:embedding}
 \end{equation}
where $e(\cdot)$ extract features for a single image. In our experiment, we use a ImageNet pretrained ResNet-50 to extract image features, and use the averaged output of the last convolution layer as the image embedding. Essentially, the embedding for each domain is the average of image feature over all the images in that domain. 

To train the network, we solve the following optimization problem with stochastic gradient descent (SGD)
\begin{equation}
    \underset{\mathcal{W}, \theta}{\text{min}} ~ \sum_{\tau }  \sum_{(x, y) \in \mathcal{D}_\tau} \mathcal{L} (f (\mathcal{W}(\mathcal{D_\tau}), x), y). 
    \label{eqn:DADN_opt}
\end{equation}
In each step of training, we first sample a domain $\mathcal{D}_\tau$ from $\{\mathcal{D}_\tau\}_{\tau=1}^T$. We compute the domain embedding vector with equation (\ref{eqn:embedding}), compute the weight factors for each layer with equation (\ref{eqn:alpha}), and compute the combined weight $\mathcal{W}(\mathcal{D}_\tau)$ with equation (\ref{eqn:dadn_weight}). Then, we sample data from the chosen domain $\mathcal{D}_\tau$ and compute the loss function in equation (\ref{eqn:DADN_opt}). We then use stochastic gradient descent to update the network weights $W_{i,j}$ in each layer as well as the controller parameters $\theta_{i,j}$. Optionally, we can also train the domain embedding vector from equation (\ref{eqn:embedding}) by propagating the gradient to the feature extraction networks. In our experiment, however, we keep domain embedding vectors fixed to reduce the computational cost of training. 

\begin{algorithm}
\SetAlgoLined
\KwIn{Training data divided in domains $\{\mathcal{D}_\tau\}_{\tau=1}^T$}
\KwOut{Optimized network weights $\{W_{i,j}^*\}$ and controller weights $\{\theta_{i,j}^*\}$}
 \For{ step $m$ }{
  Sample $\mathcal{D}_\tau \in \{\mathcal{D}_\tau\}_{\tau=1}^T$ \;
  Compute $W(\mathcal{D}_\tau)$ with equation (\ref{eqn:dadn_weight})(\ref{eqn:alpha})(\ref{eqn:embedding}) \;
  Sample a batch of $(x, y) \in \mathcal{D}_\tau$ \;
  Compute the loss function in equation (\ref{eqn:DADN_opt})\;
  Gradient update on $\{W_{i,j}\}, \{\theta_{i,j}\}$
  }
 \caption{Training a domain-aware dynamic network}
 \label{alg:dadn_train}
\end{algorithm}

Note that during training, we do need to explicitly divide a dataset into domains. The partition can be based on arbitrary attributes, but in order to utilize the data locality, we should choose attributes like weather conditions, geo-locations, and time-of-the-day, since these attributes are relatively stable during some period of time. 

During the inference, if we explicitly know the attributes of the environment, we can directly use the domain embedding corresponding to the environment to compute the network weights. However, if the attributes are not available, we can collect samples from the deployed environment to form a domain $\mathcal{D}_\tau$ and then compute DDN's weights $W(\mathcal{D}_\tau)$.  

Since each layer of DDN contains multiple weights, the capacity of the network becomes larger. On the other hand, when deployed to a given domain, the weight of the network is combined from several weights to the domain embedding. As long as the input domain stays the same, the network weights are also fixed, so we can reuse the same DDN as a small and static network for all the input. This greatly reduces the inference-time computational cost.

\subsection{Comparing DDN with related methods}
\label{sec:DADN_comparison}

\textbf{Model-pool}: We compare DDN with two related methods. First, we compare it with the model-pool idea described in Section \ref{sec:model_pool}. As mentioned before, the model pool idea mainly suffers from two problems: 1) the number of models in the pool is the same as the number of domains, which grows exponentially with the number of attributes. If we need to deal with a large number of domains, it is not feasible to store all the models on the edge device. 2) After the dataset partition, although the data diversity of each domain is smaller, the number of training samples also decreases significantly. This makes it difficult to train a specialized neural network, and each specialized model might not be able to learn cross-domain invariant features. In comparison, DDN avoids the above two problems. 1) The size of DDN do not grow with the number of domains. The choice of number of weights each layer is orthogonal with the number of domains. 2) During the training of DDN, each layer's multiple weights are all trained on the entire dataset $\mathcal{D}$. This allows each specialized model $\mathcal{W}(\mathcal{D_\tau})$ to be trained on sufficient data and they can learn cross-domain invariant features. 

\textbf{Soft-conditional computation} \cite{yang2019soft}: At each layer of DDN, the weight is computed as a linear combination of several weight tensors. This idea is inspired from Soft-Conditional Computation \cite{yang2019soft}. The advantage is that during the inference, the FLOP of SCC only increases mildly in order to compute the linear combination of weights on the fly. However, SCC models have many times more parameters. And because SCC is sample-dependent, the weight of each layer needs to be re-computed for all the input samples, and all the excessive model parameters need to be loaded from off-chip memory. This can lead to significant delay and energy consumption \cite{pedram2016dark}, especially for edge processors with limited on-chip memory. In comparison, DDN leverages the data locality to construct a small model once and reuse it for all the input samples as long as the input domain does not change. On average, DDN requires no extra FLOPs nor memory accesses than a single static network.

\section{Experiments}
Since or goal is to leverage data locality, we need to find datasets with this property. This filters out a lot of common benchmarks, such as ImageNet and COCO datasets, since there are no obvious attributes to divide the domain such that the domain remains relatively stable in a short period of time. Eventually, we decided to train an object detection network on BDD100k dataset to validate our idea.

\subsection{Datasets}
BDD100k\cite{yu2018bdd100k} is a large-scale, diverse driving dataset, which contains 70000 training images, 10000 validation images and 20000 test images with bounding box annotations for 10 categories. Images in BDD100K are all collected from real vehicles as video sequences. As a result, the input data has pretty good locality. More importantly, BDD100K annotates three attributes for each image: weather condition, scenes, time of day. So, we relied on these attributes to divide the dataset into domains. 

\subsection{Performance comparison}
In our experiment, we compare DDN with two baselines: a single static network, and a pool of specialized models. 

We first build a static network for object detection as a baseline. We choose the light-weight FBNet \cite{wu2019fbnet} + Faster RCNN.  To modify it to a domain-aware dynamic network, we expand the FBNet backbone to a supernet that each of its layer contains 2 or 4 weights. The modification is only on the backbone, and the detector part remains the same. 
Following \cite{he2019rethinking}, we train FBNet Faster RCNN for 90000 iterations from scratch with batch size as 128 on BDD100K road object detection. The initial learning rate is 0.12 and is decreased by 10 times after 60000 and 80000 iterations. 
The popular image resolution for mobile device of short edge being 320 is adopted in this group of experiments. We adopt two popular data augmentations, color jitter with strength as $0.1$ and multi-scale jitter with scales sampled from [0.75, 0.875, 1.0, 1.125, 1.25].

To partition the domains, we divide the training data into 25 domains using  two image attributes \textit{Weather condition} and \textit{Time of day}. The maximum number of images for each domain, the median, and the minimum number is 22884, 527, 2, respectively. 
To construct the model pool, we first train a network on the entire dataset, then use the weights to initialize specialized models and fine-tune them on each domains. To determine the training hyper-parameters, we used grid-search to tune the hyper parameter on one domain and share the same hyper-parameters when finetuning on other domains. This is a compromise to avoid the computational cost of tuning on each domain. For domains with less than 16 images, we do not finetune on them, but instead use the original model trained on the entire dataset. Finally, we train the domain-aware dynamic networks using the same hyper-parameters as training a single network. 

We report the AP50 (average precision with IOU $>$ 0.50) on the BDD100K benchmark in Table~\ref{tab:compare-1}. Results in Table~\ref{tab:compare-1} suggest that increasing model parameters does lead to monotonically increasing performance. Notice that during the inference, the model size of DDN is the same as a static network, since as long as the input domain stays the same, the same network can be re-used. 

\begin{table}[h]
\begin{center}
\begin{tabular}{c|ccc}
\hline
           & \begin{tabular}[c]{@{}c@{}}Total\\ Params (M)\end{tabular} & \begin{tabular}[c]{@{}c@{}}Inference\\ Params (M)\end{tabular} & AP50 (\%)          \\ \hline
Static net & 1.7                                                        & 1.7                                                            & 37.4          \\
Model pool & 43.1                                                       & 1.7                                                            & 37.1          \\
DDN 2x    & 2.8                                                        & 1.7                                                            & 39.6          \\
DDN 4x    & 4.0                                                        & 1.7                                                            & \textbf{40.0} \\ \hline
\end{tabular}
\end{center}
\caption{BDD100K validation results for DDN with FBNet+Faster-RCNN. The input image resized to ensure short edge is 320. The coloum-``Total Params'' indicates the total number of parameters. Note that at inference stage, DDN only need to store the generated static network in memory, so it has the same memory cost as the static network baseline. DDN 2x, DDN 4x represents expanding the baseline static FBNet backbone by 2 and 4 times respectively. } 
\label{tab:compare-1}
\end{table}

One surprising observation in Table~\ref{tab:compare-1} is that the a pool of specialized model does not perform better than a static network. As discussed in Section \ref{sec:DADN_comparison}, this is likely to be caused by 1) insufficient data in each domain and 2) the model finetuned to each domain cannot learn features that are invariant across domains. 

Models in Table \ref{tab:compare-1} are trained with a relatively low input resolution in order to save computational cost. Therefore, the accuracy of all three models are relatively low. We increase the input resolution to 540 (short edge is ensured to be 540) and compare the accuracy of DDN against the static network baseline. The result is in Table \ref{tab:compare-2-larger-resolution}. We can see that with a higher resolution, the accuracy of both models increases, but DDN still out-performs the baseline. 

To study whether the granularity of domain division can make a difference to the performance of DDN, we tried three different ways to divide the data and the result is reported in Table \ref{tab:compare-2-domains}. Respectively, we divide data by time-of-the day, time $\times$ weather, or times $\times$ weather $\times$ scene. DDN trained under these divisions show minor differences in accuracy, but they all out performs the static network baseline by 1.5\% of AP50.

\begin{table}[h]
\begin{center}
\begin{tabular}{c|c}
\hline
Setting       &  AP50 (\%)    \\ \hline

Static network        & 47.8  \\ 

DDN 2x       &  49.3  \\
\hline
\end{tabular}
\end{center}
\vspace{-0.1cm}
\caption{validation results for our DDN vs. Static network with a higher image resolution of 540. } 
\label{tab:compare-2-larger-resolution}
\end{table}

In Table \ref{tab:compare-2-domains}, we show the results of domain-aware dynamic net under different domain partition strategies. With different partition strategies, DNN achieves similar accuracy, and all of them outperforms the baseline. 

\begin{table}[h]
\begin{center}
\begin{tabular}{ccc}
\hline
Setting    & Domain division     &  AP50 (\%)        \\ \hline
Static net        & - & 37.4    \\ 
DDN 2x         &  Time &  39.3  \\ 
DDN 4x         &  Time  &  39.7 \\ 

DDN 2x         &  Time  \& Weather &  39.6 \\ 
DDN 4x         &  Time \& Weather&  \textbf{40.0}   \\ 

DDN 2x         &  Time \& Weather \& Scene &  39.4  \\ 
DDN 4x         &  Time  \& Weather \& Scene &  39.6 \\ \hline
\end{tabular}
\end{center}
\caption{BDD100K validation results for DDN under different domain split strategies. The column-``Domain division'' indicates the attributes used to divide the dataset. For example, under the strategy of `Time $\times$ Weather", images taken at the same time of the day and under the same weather conditions are grouped into the same domain.} 
\label{tab:compare-2-domains}
\end{table}

\subsection{Domain-dependent vs. sample-dependent}
As we apply more and more attributes to divide the data into domains, one extreme strategy is to treat each image as a single domain, and the network eventually becomes sample-dependent, rather than domain-dependent. In this experiment, we compare two dynamism in terms of efficiency and accuracy. 

We implement a Sample-dependent Dynamic Network (SDN) based on our proposed DDN. We achieve this by 
changing the training procedure of DDN to make it sample-dependent. specifically, instead of letting each layer's controller take a domain embedding as input, we feed in image embedding vectors and compute the a different set of weight factors per image. During the inference, we feed an input image to a feature extractor and use the image embedding to calculate the weight factors on-the-fly. 
 
As we discussed in Section \ref{sec:DADN_comparison}, sample-dependent dynamic network is not hardware friendly, especially for edge devices with limited memory. To test this, we implemented DDN and SDN in Caffe2\footnote{\url{https://caffe2.ai}} and deploy them on a Macbook laptop and a Rasberry Pi 3B+ micro computer. We measure two models' speed on those devices. The Macbook is equipped with a 3.5GHZ Intel i7 CPU. The Raspberry Pi 3B+ is equipped with a Braodcom BCM2837B0 quad-core A53 CPU. For simplicity, we adopt MobileNet-V2-1x as the baseline, and use 224x224 as the input resolution and set the batch size to 1. In our measurement, we ignore SDN's cost to extract image embedding vectors. We report the efficiency measurement in Table~\ref{tab:compare-3-runtime}. We can see that although SDN's FLOPs is only 3\% higher, the actual speed is 20-30\% slower, and the slow-down is more obvious on Raspberry Pi whose memory is more limited. 




\begin{table}[h]
\begin{center}
\begin{tabular}{cccc}
\hline
  &Hardware    &  Frame/sec & FLOPs (M)          \\ \hline
DDN        & Mac & 22.96  & 301   \\ 

SDN 2x         &  Mac &  19.10 & 305   \\ 
SDN 4x        &  Mac &  19.01  & 310 \\ 
\hline 
DDN       &  RPI &  1.54  & 301  \\ 
SDN 2x         &  RPI &  1.27 & 305  \\ 
SDN 4x          &  RPI &  1.20 & 310   \\ 
\hline
\end{tabular}
\end{center}
\caption{Runtime speed of domain-depedent vs. sample-dependent dynamic networks. 
Note that the time cost for SDN does not include feature extraction. The column-\textbf{FLOPs} represents inference computational cost. In the hardware column, Mac denotes the Macbook laptop and RPI denotes the Raspberry Pi micro computer. } 
\label{tab:compare-3-runtime}
\end{table}

The accuracy of DDN vs. SDN is summarized in Table~\ref{tab:compare-2-data-dependent}. We can see that even SDN have more flexibility to re-compute the model weight for each input sample, the validation accuracy of SDN is worse than DDN. To further examine the training accuracy, we find that SDN achieved better training accuracy. The result indicates that DDN is less prone to overfitting.


\begin{table}[h]
\begin{center}
\begin{tabular}{ccc}
\hline
Setting       &  \begin{tabular}[c]{@{}c@{}}AP50\\ (Val) \end{tabular} & \begin{tabular}[c]{@{}c@{}}AP50\\ (Train) \end{tabular}       \\ \hline
Static Net    & 37.4     & 41.4    \\ \hline
SDN 2x       &  39.3   & 46.2  \\ 
SDN 4x       &  38.5    & \textbf{47.7}   \\ \hline
DDN 2x       &  39.6  & 44.7 \\ 
DDN 4x       &  \textbf{40.0}  & 45.2 \\ \hline
\end{tabular}
\end{center}
\caption{DDDN indicates data dependent version of our domain-aware dynamic networks. } 
\label{tab:compare-2-data-dependent}
\end{table}

\section{Analysis}
In this section, we continue to carry more experiments to help us understand how a domain-aware dynamic network works. 

\subsection{Domain shuffle}
We assumed that given the target domain, a controller transforms the supernet into a smaller network specialized to the target domain. To verify this, we intentionally feed DDN an embedding vector from another domain and observe the performance change. In our experiment, we trained a DDN based with domains divided by time of day and weather. During evaluation, instead of feeding in the correct domain embedding extracted from the validation data, we randomly shuffle the domain embedding vectors and feed them into the network. We repeat this for 20 times and report the average and worst accuracy. From Table ~\ref{tab:ablation-shuffle}, we can see a clear accuracy drop, indicating that DDN is domain-aware. 

\begin{table}[h]
\begin{center}

\begin{tabular}{c|ccc}
\hline
        &  Normal & Worst case & Average       \\ \hline
Static network    & 37.4     & -  & -   \\
DDN 2x&  39.6   & 35.3    & 36.5 \\ 
DDN 4x&  40.0   & 34.2    &  36.2 \\ \hline

\end{tabular}
\end{center}
\caption{Results for domain shuffle. The column-\textbf{Normal} represents the performance when the correct domain embedding vectors are provided to the network. } 
\label{tab:ablation-shuffle}
\end{table}

\subsection{Mismatch domain split strategies}
Consider a DDN that is trained under one domain-split strategy, for example, split by time-of-the-day. The question is, during the inference, if we split the domain by a different strategy such as weather, can the network adapt to the new split strategy? 

To test this, we trained two DDNs, one under the split strategy of ``time'' $\times$ ``weather'' and the other under the strategy of only ``time''. During the evaluation, however, we group domains by ``scenes'', and feed domain vectors to the network. The result is reported in Table \ref{tab:ablation-mixmatch}. We can see that when the evaluation and training strategies mismatch, the accuracy drops. This indicates that since the controller does not generalize to a new split strategy. To provide another reference, we treat the entire dataset as one domain and compute the embedding vector. We then use this embedding vector to feed into two networks and evlaute the accuracy. The accuracy is almost the same as the case with mismatched splitting strategies.  We conjecture that this is because the controller is not sensitive to attributes that are used to divide new domains. Therefore, it failed to find a network specialized to the target domain. Instead, the controller find a generalized network that is optimized to the entire dataset. Also note that even under mismatched domain split, the DDN still outperforms the baseline of a single static network. 





\begin{table}[h]
\begin{center}
\begin{tabular}{c|ccc}
\hline
    Training split     &  Matched   & Mismatched & Reference \\ \hline
T+W &  39.6   & 38.52 & 38.54  \\ 
T &  39.3   & 38.08 & 38.05 \\ \hline 
\end{tabular}
Accuracy of a static network: 37.4
\end{center}
\caption{Results of mismatched domain split. We trained two DDNs under two domain split strategy, one divided by time and weather (row-``T+W''), and another by time only (row-``T''). Column-''Matched'' denotes the accuracy of the network if the domain split matches with the training splits. Column-``Mismatched'' denotes the case where we split the evaluation dataset by ``scene''. Reference denotes the accuracy when we feed entire dataset's embedding vector to the two networks.
} 
\label{tab:ablation-mixmatch}
\end{table}





\subsection{Visualization of the weight factors}
In this subsection, we visualize the weight factors. We split domains based on ``time of day" $\times$ ``weather", and we choose 4 domains to visualize: ``clear-daytime'', ``clear-night'', ``rainy-daytime'', ``rainy-night''. 

\begin{figure}[ht]
\centering
\includegraphics[width=0.9 \linewidth]{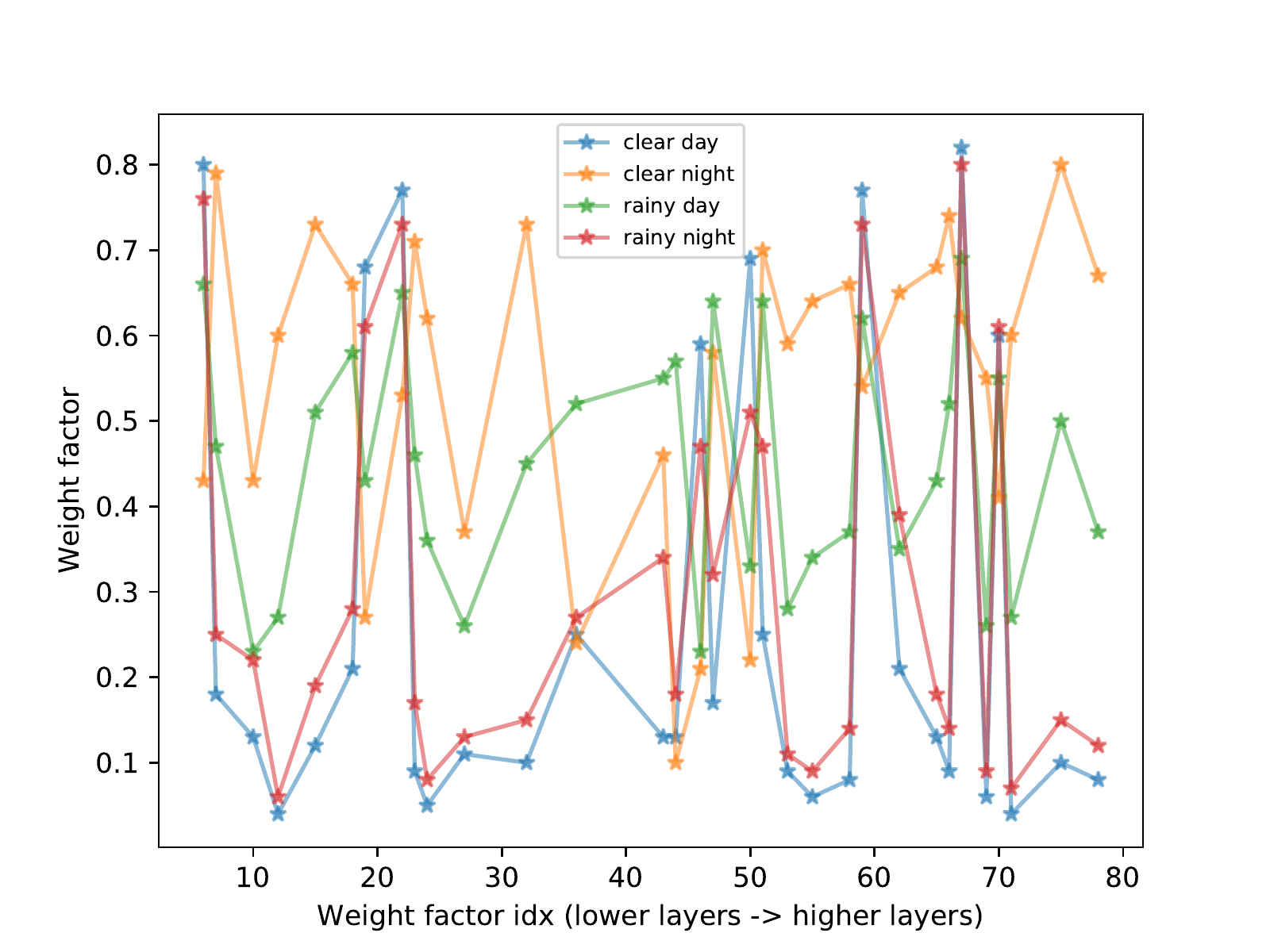}
\caption{Weight factor visualization for four domains: ``clear-daytime'', ``clear-night'', ``rainy-daytime'', and ``rainy-night''. }
\label{fig:visual-routing-weights-domain}
\end{figure}

For each specialized network, we concatenate the weight factors of all layers from start to end into a 80-dimension vector, and we plot the vector in Figure ~\ref{fig:visual-routing-weights-domain}. For clearer visualization, we only plot 32 weights with the highest variance. Some interesting findings include: 1) For the same time of day, weight factors for different weather conditions are quite different. Similarly, with the same weather condition, weight factors for different daytime can also be quite different. 2) The difference of weight factors do not always follow human intuitions. For example, weight factors for ``clear-daytime'' (blue curve) is very similar to that of ``rainy night'' (red curve), even though both weather and time conditions are different.

\begin{figure}[ht]
\centering

\includegraphics[width=0.9\linewidth]{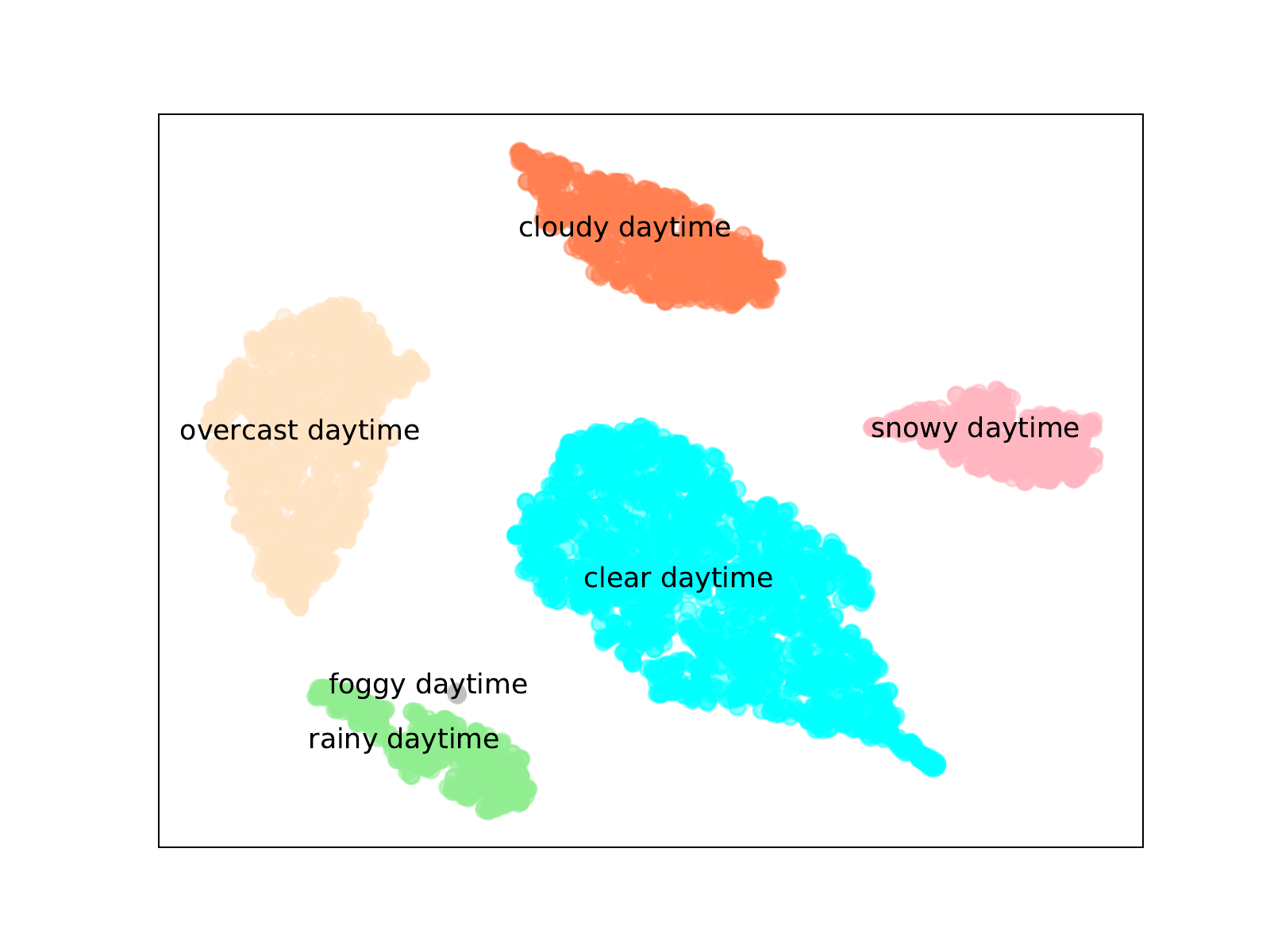}   \\

\caption{t-SNE visualization of  weight factors for different input images on the validation set. }
\label{fig:visual-routing-weights-data}
\end{figure}

\subsection{Data embedding visualization}
Following the previous subsection, for a given domain, we can use a weight factor vector to represent a domain. In fact, for each image, we can also compute the weight factors and use it to represent an image. Here, we visualize each image using its corresponding weigh factor vector. To reduce the dimension, we use t-SNE\cite{maaten2008visualizing, van2014accelerating}, which is particularly well suited for the visualization of high-dimensional datasets. We show the visualization results for images taken in ``clear-daytime'', ``rainy-daytime'', ``snowy-daytime'', ``cloudy-daytime'', ``foggy-daytime'', and ``overcast-daytime'' in Figure~\ref{fig:visual-routing-weights-data}. From the visualization, it is very clear that images from the same domain are projected in clusters. This verifies that the controller tend to pick similar sub-networks from a DDN to process images from the same domain. 

\section{Conclusion}
In this paper, we propose Domain-aware Dynamic Networks (DDN), a novel framework to improve a deep neural network's capacity without increasing its inference-time computational cost. DDN is a high-capacity dynamic network with each layer having multiple weights. Given an input domain, DDN dynamically combines weights and generates a low complexity model specialized in the input domain. Thanks to the data locality, for an edge device and within a short period of time, the domain of input data to the edge device remains relatively stable. This allows us to reuse the generated low-complexity model as long as the input domain does not change. Our experiments on the BDD100K benchmark shows that without increasing any inference-time parameters, FLOPs, and actual latency, a DDN can achieve up to 2.6\% higher AP50 compared with static networks and other baseline methods. 

{\small
\bibliographystyle{ieee_fullname}
\bibliography{egbib}

\begin{thebibliography}{10}\itemsep=-1pt

\bibitem{al2017continuous}
Maruan Al-Shedivat, Trapit Bansal, Yuri Burda, Ilya Sutskever, Igor Mordatch,
  and Pieter Abbeel.
\newblock Continuous adaptation via meta-learning in nonstationary and
  competitive environments.
\newblock {\em arXiv preprint arXiv:1710.03641}, 2017.

\bibitem{bejnordi2019batch}
Babak~Ehteshami Bejnordi, Tijmen Blankevoort, and Max Welling.
\newblock Batch-shaped channel gated networks.
\newblock {\em arXiv preprint arXiv:1907.06627}, 2019.

\bibitem{cai2019once}
Han Cai, Chuang Gan, and Song Han.
\newblock Once for all: Train one network and specialize it for efficient
  deployment.
\newblock {\em arXiv preprint arXiv:1908.09791}, 2019.

\bibitem{dai2017deformable}
Jifeng Dai, Haozhi Qi, Yuwen Xiong, Yi Li, Guodong Zhang, Han Hu, and Yichen
  Wei.
\newblock Deformable convolutional networks.
\newblock In {\em Proceedings of the IEEE international conference on computer
  vision}, pages 764--773, 2017.

\bibitem{dai2019chamnet}
Xiaoliang Dai, Peizhao Zhang, Bichen Wu, Hongxu Yin, Fei Sun, Yanghan Wang,
  Marat Dukhan, Yunqing Hu, Yiming Wu, Yangqing Jia, et~al.
\newblock Chamnet: Towards efficient network design through platform-aware
  model adaptation.
\newblock In {\em Proceedings of the IEEE Conference on Computer Vision and
  Pattern Recognition}, pages 11398--11407, 2019.

\bibitem{ding2019approximated}
Xiaohan Ding, Guiguang Ding, Yuchen Guo, Jungong Han, and Chenggang Yan.
\newblock Approximated oracle filter pruning for destructive cnn width
  optimization.
\newblock {\em arXiv preprint arXiv:1905.04748}, 2019.

\bibitem{dong2019hawq}
Zhen Dong, Zhewei Yao, Amir Gholami, Michael Mahoney, and Kurt Keutzer.
\newblock Hawq: Hessian aware quantization of neural networks with
  mixed-precision.
\newblock {\em arXiv preprint arXiv:1905.03696}, 2019.

\bibitem{finn2017model}
Chelsea Finn, Pieter Abbeel, and Sergey Levine.
\newblock Model-agnostic meta-learning for fast adaptation of deep networks.
\newblock In {\em Proceedings of the 34th International Conference on Machine
  Learning-Volume 70}, pages 1126--1135. JMLR. org, 2017.

\bibitem{gholami2018squeezenext}
Amir Gholami, Kiseok Kwon, Bichen Wu, Zizheng Tai, Xiangyu Yue, Peter Jin,
  Sicheng Zhao, and Kurt Keutzer.
\newblock Squeezenext: Hardware-aware neural network design.
\newblock In {\em Proceedings of the IEEE Conference on Computer Vision and
  Pattern Recognition Workshops}, pages 1638--1647, 2018.

\bibitem{guo2019single}
Zichao Guo, Xiangyu Zhang, Haoyuan Mu, Wen Heng, Zechun Liu, Yichen Wei, and
  Jian Sun.
\newblock Single path one-shot neural architecture search with uniform
  sampling.
\newblock {\em arXiv preprint arXiv:1904.00420}, 2019.

\bibitem{han2015deep}
Song Han, Huizi Mao, and William~J Dally.
\newblock Deep compression: Compressing deep neural networks with pruning,
  trained quantization and huffman coding.
\newblock {\em arXiv preprint arXiv:1510.00149}, 2015.

\bibitem{he2019rethinking}
Kaiming He, Ross Girshick, and Piotr Doll{\'a}r.
\newblock Rethinking imagenet pre-training.
\newblock In {\em Proceedings of the IEEE International Conference on Computer
  Vision}, pages 4918--4927, 2019.

\bibitem{he2017channel}
Yihui He, Xiangyu Zhang, and Jian Sun.
\newblock Channel pruning for accelerating very deep neural networks.
\newblock In {\em Proceedings of the IEEE International Conference on Computer
  Vision}, pages 1389--1397, 2017.

\bibitem{hoffman2017cycada}
Judy Hoffman, Eric Tzeng, Taesung Park, Jun-Yan Zhu, Phillip Isola, Kate
  Saenko, Alexei~A Efros, and Trevor Darrell.
\newblock Cycada: Cycle-consistent adversarial domain adaptation.
\newblock {\em arXiv preprint arXiv:1711.03213}, 2017.

\bibitem{howard2019searching}
Andrew Howard, Mark Sandler, Grace Chu, Liang-Chieh Chen, Bo Chen, Mingxing
  Tan, Weijun Wang, Yukun Zhu, Ruoming Pang, Vijay Vasudevan, et~al.
\newblock Searching for mobilenetv3.
\newblock {\em arXiv preprint arXiv:1905.02244}, 2019.

\bibitem{howard2017mobilenets}
Andrew~G Howard, Menglong Zhu, Bo Chen, Dmitry Kalenichenko, Weijun Wang,
  Tobias Weyand, Marco Andreetto, and Hartwig Adam.
\newblock Mobilenets: Efficient convolutional neural networks for mobile vision
  applications.
\newblock {\em arXiv preprint arXiv:1704.04861}, 2017.

\bibitem{jaderberg2015spatial}
Max Jaderberg, Karen Simonyan, Andrew Zisserman, et~al.
\newblock Spatial transformer networks.
\newblock In {\em Advances in neural information processing systems}, pages
  2017--2025, 2015.

\bibitem{li2017learning}
Zhizhong Li and Derek Hoiem.
\newblock Learning without forgetting.
\newblock {\em IEEE transactions on pattern analysis and machine intelligence},
  40(12):2935--2947, 2017.

\bibitem{lin2017runtime}
Ji Lin, Yongming Rao, Jiwen Lu, and Jie Zhou.
\newblock Runtime neural pruning.
\newblock In {\em Advances in Neural Information Processing Systems}, pages
  2181--2191, 2017.

\bibitem{liu2019metapruning}
Zechun Liu, Haoyuan Mu, Xiangyu Zhang, Zichao Guo, Xin Yang, Tim Kwang-Ting
  Cheng, and Jian Sun.
\newblock Metapruning: Meta learning for automatic neural network channel
  pruning.
\newblock {\em arXiv preprint arXiv:1903.10258}, 2019.

\bibitem{ma2018shufflenet}
Ningning Ma, Xiangyu Zhang, Hai-Tao Zheng, and Jian Sun.
\newblock Shufflenet v2: Practical guidelines for efficient cnn architecture
  design.
\newblock In {\em Proceedings of the European Conference on Computer Vision
  (ECCV)}, pages 116--131, 2018.

\bibitem{maaten2008visualizing}
Laurens van~der Maaten and Geoffrey Hinton.
\newblock Visualizing data using t-sne.
\newblock {\em Journal of machine learning research}, 9(Nov):2579--2605, 2008.

\bibitem{mahajan2018exploring}
Dhruv Mahajan, Ross Girshick, Vignesh Ramanathan, Kaiming He, Manohar Paluri,
  Yixuan Li, Ashwin Bharambe, and Laurens van~der Maaten.
\newblock Exploring the limits of weakly supervised pretraining.
\newblock In {\em Proceedings of the European Conference on Computer Vision
  (ECCV)}, pages 181--196, 2018.

\bibitem{pedram2016dark}
Ardavan Pedram, Stephen Richardson, Mark Horowitz, Sameh Galal, and Shahar
  Kvatinsky.
\newblock Dark memory and accelerator-rich system optimization in the dark
  silicon era.
\newblock {\em IEEE Design \& Test}, 34(2):39--50, 2016.

\bibitem{recht2019imagenet}
Benjamin Recht, Rebecca Roelofs, Ludwig Schmidt, and Vaishaal Shankar.
\newblock Do imagenet classifiers generalize to imagenet?
\newblock {\em arXiv preprint arXiv:1902.10811}, 2019.

\bibitem{sandler2018mobilenetv2}
Mark Sandler, Andrew Howard, Menglong Zhu, Andrey Zhmoginov, and Liang-Chieh
  Chen.
\newblock Mobilenetv2: Inverted residuals and linear bottlenecks.
\newblock In {\em Proceedings of the IEEE Conference on Computer Vision and
  Pattern Recognition}, pages 4510--4520, 2018.

\bibitem{shao2019objects365}
Shuai Shao, Zeming Li, Tianyuan Zhang, Chao Peng, Gang Yu, Xiangyu Zhang, Jing
  Li, and Jian Sun.
\newblock Objects365: A large-scale, high-quality dataset for object detection.
\newblock In {\em Proceedings of the IEEE International Conference on Computer
  Vision}, pages 8430--8439, 2019.

\bibitem{shazeer2017outrageously}
Noam Shazeer, Azalia Mirhoseini, Krzysztof Maziarz, Andy Davis, Quoc Le,
  Geoffrey Hinton, and Jeff Dean.
\newblock Outrageously large neural networks: The sparsely-gated
  mixture-of-experts layer.
\newblock {\em arXiv preprint arXiv:1701.06538}, 2017.

\bibitem{sun2019test}
Yu Sun, Xiaolong Wang, Zhuang Liu, John Miller, Alexei~A Efros, and Moritz
  Hardt.
\newblock Test-time training for out-of-distribution generalization.
\newblock {\em arXiv preprint arXiv:1909.13231}, 2019.

\bibitem{tan2019mnasnet}
Mingxing Tan, Bo Chen, Ruoming Pang, Vijay Vasudevan, Mark Sandler, Andrew
  Howard, and Quoc~V Le.
\newblock Mnasnet: Platform-aware neural architecture search for mobile.
\newblock In {\em Proceedings of the IEEE Conference on Computer Vision and
  Pattern Recognition}, pages 2820--2828, 2019.

\bibitem{van2014accelerating}
Laurens Van Der~Maaten.
\newblock Accelerating t-sne using tree-based algorithms.
\newblock {\em The Journal of Machine Learning Research}, 15(1):3221--3245,
  2014.

\bibitem{veit2018convolutional}
Andreas Veit and Serge Belongie.
\newblock Convolutional networks with adaptive inference graphs.
\newblock In {\em Proceedings of the European Conference on Computer Vision
  (ECCV)}, pages 3--18, 2018.

\bibitem{wang2019haq}
Kuan Wang, Zhijian Liu, Yujun Lin, Ji Lin, and Song Han.
\newblock Haq: Hardware-aware automated quantization with mixed precision.
\newblock In {\em Proceedings of the IEEE Conference on Computer Vision and
  Pattern Recognition}, pages 8612--8620, 2019.

\bibitem{wang2018skipnet}
Xin Wang, Fisher Yu, Zi-Yi Dou, Trevor Darrell, and Joseph~E Gonzalez.
\newblock Skipnet: Learning dynamic routing in convolutional networks.
\newblock In {\em Proceedings of the European Conference on Computer Vision
  (ECCV)}, pages 409--424, 2018.

\bibitem{wang2019tafe}
Xin Wang, Fisher Yu, Ruth Wang, Trevor Darrell, and Joseph~E Gonzalez.
\newblock Tafe-net: Task-aware feature embeddings for low shot learning.
\newblock In {\em Proceedings of the IEEE Conference on Computer Vision and
  Pattern Recognition}, pages 1831--1840, 2019.

\bibitem{Wu:EECS-2019-120}
Bichen Wu.
\newblock {\em Efficient Deep Neural Networks}.
\newblock PhD thesis, EECS Department, University of California, Berkeley, Aug
  2019.

\bibitem{wu2019fbnet}
Bichen Wu, Xiaoliang Dai, Peizhao Zhang, Yanghan Wang, Fei Sun, Yiming Wu,
  Yuandong Tian, Peter Vajda, Yangqing Jia, and Kurt Keutzer.
\newblock Fbnet: Hardware-aware efficient convnet design via differentiable
  neural architecture search.
\newblock In {\em Proceedings of the IEEE Conference on Computer Vision and
  Pattern Recognition}, pages 10734--10742, 2019.

\bibitem{wu2017squeezedet}
Bichen Wu, Forrest Iandola, Peter~H Jin, and Kurt Keutzer.
\newblock Squeezedet: Unified, small, low power fully convolutional neural
  networks for real-time object detection for autonomous driving.
\newblock In {\em Proceedings of the IEEE Conference on Computer Vision and
  Pattern Recognition Workshops}, pages 129--137, 2017.

\bibitem{wu2018shift}
Bichen Wu, Alvin Wan, Xiangyu Yue, Peter Jin, Sicheng Zhao, Noah Golmant, Amir
  Gholaminejad, Joseph Gonzalez, and Kurt Keutzer.
\newblock Shift: A zero flop, zero parameter alternative to spatial
  convolutions.
\newblock In {\em Proceedings of the IEEE Conference on Computer Vision and
  Pattern Recognition}, pages 9127--9135, 2018.

\bibitem{wu2018squeezeseg}
Bichen Wu, Alvin Wan, Xiangyu Yue, and Kurt Keutzer.
\newblock Squeezeseg: Convolutional neural nets with recurrent crf for
  real-time road-object segmentation from 3d lidar point cloud.
\newblock In {\em 2018 IEEE International Conference on Robotics and Automation
  (ICRA)}, pages 1887--1893. IEEE, 2018.

\bibitem{wu2018mixed}
Bichen Wu, Yanghan Wang, Peizhao Zhang, Yuandong Tian, Peter Vajda, and Kurt
  Keutzer.
\newblock Mixed precision quantization of convnets via differentiable neural
  architecture search.
\newblock {\em arXiv preprint arXiv:1812.00090}, 2018.

\bibitem{wu2019squeezesegv2}
Bichen Wu, Xuanyu Zhou, Sicheng Zhao, Xiangyu Yue, and Kurt Keutzer.
\newblock Squeezesegv2: Improved model structure and unsupervised domain
  adaptation for road-object segmentation from a lidar point cloud.
\newblock In {\em 2019 International Conference on Robotics and Automation
  (ICRA)}, pages 4376--4382. IEEE, 2019.

\bibitem{yang2019soft}
Brandon Yang, Gabriel Bender, Quoc~V Le, and Jiquan Ngiam.
\newblock Soft conditional computation.
\newblock {\em arXiv preprint arXiv:1904.04971}, 2019.

\bibitem{yang2019synetgy}
Yifan Yang, Qijing Huang, Bichen Wu, Tianjun Zhang, Liang Ma, Giulio
  Gambardella, Michaela Blott, Luciano Lavagno, Kees Vissers, John Wawrzynek,
  et~al.
\newblock Synetgy: Algorithm-hardware co-design for convnet accelerators on
  embedded fpgas.
\newblock In {\em Proceedings of the 2019 ACM/SIGDA International Symposium on
  Field-Programmable Gate Arrays}, pages 23--32. ACM, 2019.

\bibitem{yu2018bdd100k}
Fisher Yu, Wenqi Xian, Yingying Chen, Fangchen Liu, Mike Liao, Vashisht
  Madhavan, and Trevor Darrell.
\newblock Bdd100k: A diverse driving video database with scalable annotation
  tooling.
\newblock {\em arXiv preprint arXiv:1805.04687}, 2018.

\bibitem{yue2019domain}
Xiangyu Yue, Yang Zhang, Sicheng Zhao, Alberto Sangiovanni-Vincentelli, Kurt
  Keutzer, and Boqing Gong.
\newblock Domain randomization and pyramid consistency: Simulation-to-real
  generalization without accessing target domain data.
\newblock In {\em Proceedings of the IEEE International Conference on Computer
  Vision}, pages 2100--2110, 2019.

\bibitem{zhang2018shufflenet}
Xiangyu Zhang, Xinyu Zhou, Mengxiao Lin, and Jian Sun.
\newblock Shufflenet: An extremely efficient convolutional neural network for
  mobile devices.
\newblock In {\em Proceedings of the IEEE Conference on Computer Vision and
  Pattern Recognition}, pages 6848--6856, 2018.

\bibitem{zhao2019multi}
Sicheng Zhao, Bo Li, Xiangyu Yue, Yang Gu, Pengfei Xu, Runbo Hu, Hua Chai, and
  Kurt Keutzer.
\newblock Multi-source domain adaptation for semantic segmentation.
\newblock {\em arXiv preprint arXiv:1910.12181}, 2019.

\bibitem{zhao2018unsupervised}
Sicheng Zhao, Bichen Wu, Joseph Gonzalez, Sanjit~A Seshia, and Kurt Keutzer.
\newblock Unsupervised domain adaptation: from simulation engine to the
  realworld.
\newblock {\em arXiv preprint arXiv:1803.09180}, 2018.

\bibitem{zhu2019deformable}
Xizhou Zhu, Han Hu, Stephen Lin, and Jifeng Dai.
\newblock Deformable convnets v2: More deformable, better results.
\newblock In {\em Proceedings of the IEEE Conference on Computer Vision and
  Pattern Recognition}, pages 9308--9316, 2019.

\bibitem{zoph2016neural}
Barret Zoph and Quoc~V Le.
\newblock Neural architecture search with reinforcement learning.
\newblock {\em arXiv preprint arXiv:1611.01578}, 2016.

\end{thebibliography}
}

\end{document}